\documentclass[conference]{IEEEtran}
\IEEEoverridecommandlockouts

\usepackage{amsmath,amsfonts,amssymb,amsthm}
\usepackage{graphicx}
\usepackage{algorithm}
\usepackage{algpseudocode}
\usepackage{booktabs}
\usepackage{url}
\usepackage{cite}
\usepackage{tikz}
\usepackage{microtype}
\usepackage[percent]{overpic}
\usetikzlibrary{calc, shapes, decorations.pathreplacing}
\usepackage{hyperref}
\hypersetup{colorlinks=true, linkcolor=blue, citecolor=blue, urlcolor=blue}

\title{SynthCharge: An Electric Vehicle Routing Instance Generator with Feasibility Screening to Enable Learning-Based Optimization and Benchmarking}

\author{
\IEEEauthorblockN{
Mertcan Daysalilar\IEEEauthorrefmark{1},
Fuat Uyguroglu\IEEEauthorrefmark{2},
Gabriel Nicolosi\IEEEauthorrefmark{3}, and
Adam Meyers\IEEEauthorrefmark{1}
}

\IEEEauthorblockA{
\IEEEauthorrefmark{1}Industrial and Systems Engineering, University of Miami, Coral Gables, FL, USA
}

\IEEEauthorblockA{
\IEEEauthorrefmark{2}Faculty of Engineering, Cyprus International University 99258 Nicosia, North Cyprus, via Mersin 10, Turkey
}

\IEEEauthorblockA{
\IEEEauthorrefmark{3}Engineering Management and Systems Engineering, Missouri University of Science and Technology, Rolla, MO, USA
}
}

\begin{document}
\maketitle
\begin{abstract}
    The electric vehicle routing problem with time windows (EVRPTW) extends the classical VRPTW by introducing battery capacity constraints and charging station decisions. Existing benchmark datasets are often static and lack verifiable feasibility, which restricts reproducible evaluation of learning-based routing models. We introduce SynthCharge, a parametric generator that produces diverse, feasibility-screened EVRPTW instances across varying spatiotemporal configurations and scalable customer counts. While SynthCharge can currently generate large-scale instances of up to 500 customers, we focus our experiments on sizes ranging from 5 to 100 customers. Unlike static benchmark suites, SynthCharge integrates instance geometry with adaptive energy capacity scaling and range-aware charging station placement. To guarantee structural validity, the generator systematically filters out unsolvable instances through a fast feasibility screening process. Ultimately, SynthCharge provides the dynamic benchmarking infrastructure needed to systematically evaluate the robustness of emerging neural routing and data-driven approaches. 
    
\end{abstract}


\section{Introduction}
Electrification of urban freight and last-mile delivery makes energy-constrained routing a core problem in sustainable intelligent transportation systems (ITS). The electric vehicle routing problem with time windows (EVRPTW) extends the VRPTW by incorporating battery constraints and decisions about where and when to recharge \cite{Schneider2014,Erdogan2012}. These additions fundamentally change route construction and feasibility. While exact and hybrid optimization methods have produced effective algorithms for multiple EVRPTW variants \cite{Desaulniers2016,Hiermann2016,Keskin2016,goeke2019}, their high computational costs limit real-time application. This limitation has driven a shift toward learning-based routing methods, which offer near real-time route generation and data-driven policy construction \cite{nazari2018reinforcement,kool2019attention}. Moreover, this shift increases the need for evaluation protocols that go beyond a single finite benchmark suite and instead characterize behavior under systematic variation in instance characteristics. Related ITS research also studies EV routing and charging optimization in operational settings, including charging-routing formulations and uncertainty-aware battery-electric truck dispatching with time windows, further motivating benchmark infrastructure that supports controlled, reproducible evaluation across constraint regimes \cite{DeAndoin2023EVCRP,Peng2024RobustBET,DeNunzio2020ITSC}.

A consistent finding in recent neural routing literature is that generalization remains a primary bottleneck. Models trained on a fixed distribution of instances can degrade under shifts in spatial structure, problem size, or constraint parameters.  Studies of distribution shift and omni-generalization \cite{Jiang2023Shift,Zhou2023Omni,Luo2025Tightness} show that neural combinatorial optimization can be sensitive to structural shifts, motivating evaluation across controlled temporal and capacity regimes.  These results indicate that benchmark suites that have fixed topology and constraint regimes may be insufficient for reproducible, distribution-aware evaluation. Recent advances in learning-based EVRPTW further emphasize this sensitivity to instance regimes. Curriculum-based reinforcement learning approaches progressively expose models to increasing constraint complexity and report improvements in generalization across problem scales \cite{Daysalilar2026CB}. Such findings suggest that model performance depends strongly on how spatial structure, energy constraints, and time window tightness are distributed during training and evaluation. However, widely used EVRPTW benchmark suites provide limited mechanisms to systematically vary these structural characteristics.

Benchmark design has itself been recognized as a research direction. Early EVRPTW benchmarks were built upon the Solomon VRPTW foundation \cite{Solomon1987} and its electric extensions \cite{Schneider2014}.  While these suites remain indispensable for solver comparison, they are finite and provide limited parametric control over instance families.  More broadly, benchmark design critiques emphasize that widely adopted instance sets can become overly homogeneous with respect to structural characteristics \cite{uchoa2017}.  Community repositories such as VRP-REP improve reproducibility by consolidating instances and supporting verification workflows \cite{vrprep}, but they do not provide controllable instance generation or instance-level feasibility screening. Work on synthetic traffic generation frames the contribution around distribution-level realism and data availability rather than downstream optimization performance \cite{Nigam2023ITSC}. Complementary guidance from machine learning emphasizes dataset transparency through documentation and clear statements of intended use, limitations, and validation \cite{Gebru2021Datasheets}.  Taken together, these trends point toward a modern benchmark standard incorporating controllable generation, explicit validation, and clearly documented metadata.

To address these limitations, we introduce SynthCharge, a fully parametric EVRPTW instance generator. SynthCharge generates geometry-aware EVRPTW instances across multiple spatial topologies, time window regimes, and customer scales. Energy-related parameters (e.g., battery capacity and consumption rates) are scaled as explicit functions of instance geometry, including average inter-customer distance and depot dispersion. This geometry-conditioned scaling links energy feasibility to spatial structure, preventing trivial feasibility or systematic infeasibility as instance size or spatial spread changes. Charging station locations are placed using range-aware rules tied to vehicle travel limits rather than fixed templates.

To support explicit validation and reproducibility, each generated instance is accompanied by feasibility metadata obtained through two stages: (i) linear-time structural screening that detects obvious violations (e.g., unreachable customers under energy constraints), and (ii) optional exact mixed integer linear programming (MILP) verification for small instances ($N \le 10$). SynthCharge is positioned as a benchmark infrastructure rather than as a solver contribution; its purpose is to enable evaluation under systematically varied spatial structure, customer scale, time window tightness, and energy capacity regimes, thereby supporting distribution-aware performance assessment.

\section{Related Work}
\subsection{EVRPTW Formulations and Benchmark Suites}

Solomon's VRPTW instances \cite{Solomon1987} established canonical time window routing families with distinct spatial structures, forming the basis of decades of benchmarking practice. Schneider \emph{et al.} extended this formulation to the EVRPTW by incorporating recharging stations and battery capacity constraints \cite{Schneider2014}. Subsequent and more recent studies expanded modeling realism through partial recharging policies \cite{Keskin2016}, heterogeneous fleets and charger types \cite{Hiermann2016,goeke2019}, soft time window penalties \cite{Daysalilar2023Heterogeneous}, and exact solution methodologies \cite{Desaulniers2016}, reflecting increasing operational complexity. While these contributions enrich EVRPTW modeling and algorithmic development, the associated benchmark suites typically remain as finite instance collections. As such, they provide limited mechanisms for systematic parametric variation across spatial structure, customer scale, and energy regimes, and rarely include instance-level feasibility information that can support transparent and reproducible evaluation.

\subsection{Learning-Based Routing under Distribution Shift}
Learning-based routing methods commonly train on synthetically-generated instances sampled from fixed distributions \cite{nazari2018reinforcement,kool2019attention}.  Recent work shows that this evaluation protocol can obscure sensitivity to distribution shift.  Jiang \emph{et al.} studied vehicle routing under distribution shift \cite{Jiang2023Shift}, while Zhou \emph{et al.} formalized omni-generalization across instance size and node distribution \cite{Zhou2023Omni}.  Luo \emph{et al.} show that shifts in constraint tightness can induce systematic degradation \cite{Luo2025Tightness}.  These studies support the need for benchmarks that provide explicit dimensions of variation rather than relying on a single fixed family.

\subsection{Synthetic Benchmarks and Validation Practices in Intelligent Transportation Systems}
Recent ITS research has emphasized both structured synthetic data generation and operational optimization for electrified transportation systems. Nigam and Srivastava \cite{Nigam2023ITSC} generate synthetic traffic data and evaluate fidelity via distribution-level statistics. Complementary ITS studies address EV routing and charging optimization in application settings, including charging-routing formulations and uncertainty-aware battery-electric truck dispatching with time windows \cite{DeAndoin2023EVCRP,Peng2024RobustBET,DeNunzio2020ITSC}. However, these works do not target benchmark generation with controllable instance families and explicit instance-level feasibility screening for EVRPTW evaluation. Guidance such as “Datasheets for Datasets” emphasizes documentation, transparency of construction procedures, and clear statements of intended use and limitations \cite{Gebru2021Datasheets}.  SynthCharge follows this approach by integrating parametric instance generation with explicit feasibility screening and data reliability reporting.

\section{Methodology}
\label{sec:methodology}

This section summarizes the synthetic instance generation procedure. Table~\ref{tab:notation} lists the key notation used throughout.

\subsection{Notation and Variable Definitions}
\label{sec:notation}

We consider a single-depot EVRPTW instance defined on a node set 
$V = \{0\} \cup \mathcal{C} \cup \mathcal{S}$, where node $0$ denotes the depot, $\mathcal{C} = \{1,\dots,N\}$ denotes the set of $N$ customer indices, 
and $\mathcal{S} = \{N+1,\dots,N+|\mathcal{S}|\}$ denotes the set of external charging station indices. All node indices are positive integers. Let $d_{ij}$ denote the Euclidean distance between nodes $i$ and $j$ and $t_{ij}$ the travel time, assumed to be equal to $d_{ij}$ (unit speed). Each vehicle has load capacity $Q$ and battery capacity $B$. The energy consumption rate per unit distance is $r$, yielding a maximum travel range of $R = B / r$. Each customer $i \in \mathcal{C}$ has demand $q_i \sim \mathcal{U}(0.02Q, 0.30Q)$ and service duration $s_i \in [s_{\min}, s_{\max}]$. The planning horizon is the time interval $[0,H]$, where $H$ denotes the maximum allowable route duration. Service at customer $i$ must begin within a time window $[e_i, l_i] \subseteq [0,H]$. Time window widths are generated using a width fraction parameter $\phi \in (0,1]$, where the window width is defined as $W = \phi H$. For a given instance, $\phi$ is fixed and all customer time windows share the same width $W$, while their starting times are randomly sampled within $[0, H-W]$. In the MILP formulation used for validation, $b_i$ denotes the service start time and $y_i$ the battery energy upon arrival at node $i$. We use $\gamma$ to denote the acceptance rate, defined as the fraction of generated instances that satisfy feasibility validation relative to the total number of generated instances. All random variables are generated using an instance-specific random number generator with a fixed seed to ensure reproducibility.

\begin{table}[t]
    \caption{Key notation used in SynthCharge}
    \label{tab:notation}
    \centering
    \footnotesize
    \setlength{\tabcolsep}{4pt}
    \begin{tabular}{ll}
        \toprule
        Symbol & Meaning \\
        \midrule
        $N$ & Number of customers \\
        $\mathcal{C}$ & Set of customer indices $\{1,\dots,N\}$ \\
        $\mathcal{S}$ & Set of charging station indices \\
        $Q$ & Vehicle load capacity \\
        $B$ & Battery capacity \\
        $r$ & Energy consumption per unit distance \\
        $R$ & Maximum travel range ($R = B/r$) \\
        $d_{ij}$ & Euclidean distance between nodes $i$ and $j$ \\
        $t_{ij}$ & Travel time (assumed to be equal to $d_{ij}$) \\
        $q_i$ & Demand of customer $i$ \\
        $s_i$ & Service duration at customer $i$ \\
        $e_i, l_i$ & Earliest and latest service start times \\
        $H$ & Planning horizon upper bound (of time interval $[0,H]$) \\
        $\phi$ & Time window width fraction (common across customers) \\
        $W$ & Time window width ($W=\phi H$) \\
        $b_i$ & Service start time at node $i$ (MILP variable) \\
        $y_i$ & Battery energy upon arrival at node $i$ (MILP variable) \\
        $\gamma$ & Acceptance rate (fraction of generated instances passing validation) \\
        \bottomrule
    \end{tabular}
\end{table}

\subsection{Spatial Topology Generation}
\label{sec:spatial}
\begin{figure*}[t]
    \centering
    
    \begin{overpic}[width=0.32\linewidth]{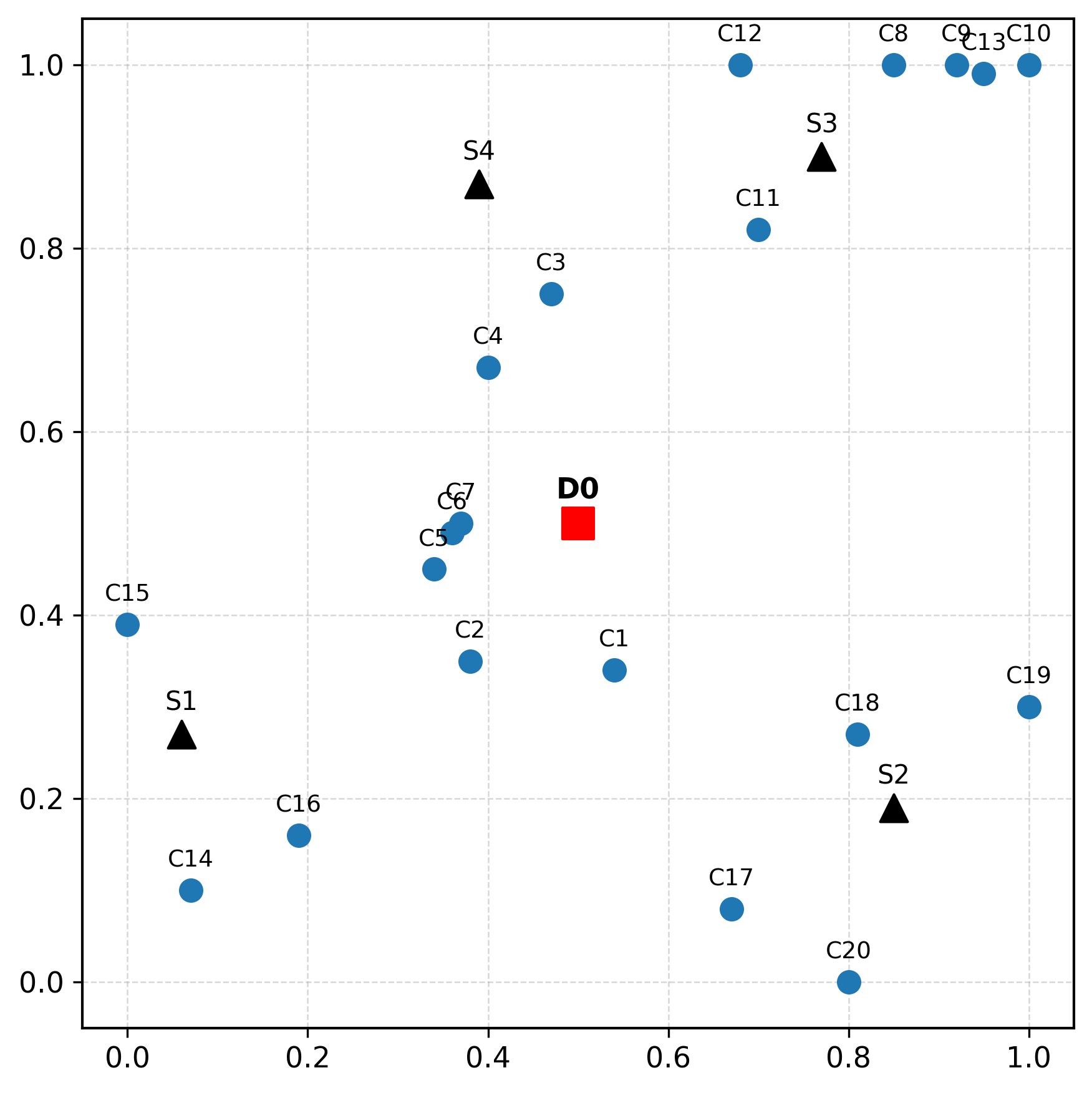}
        \put(-7,92){\textbf{(a)}}
    \end{overpic}
    \hfill
    \begin{overpic}[width=0.32\linewidth]{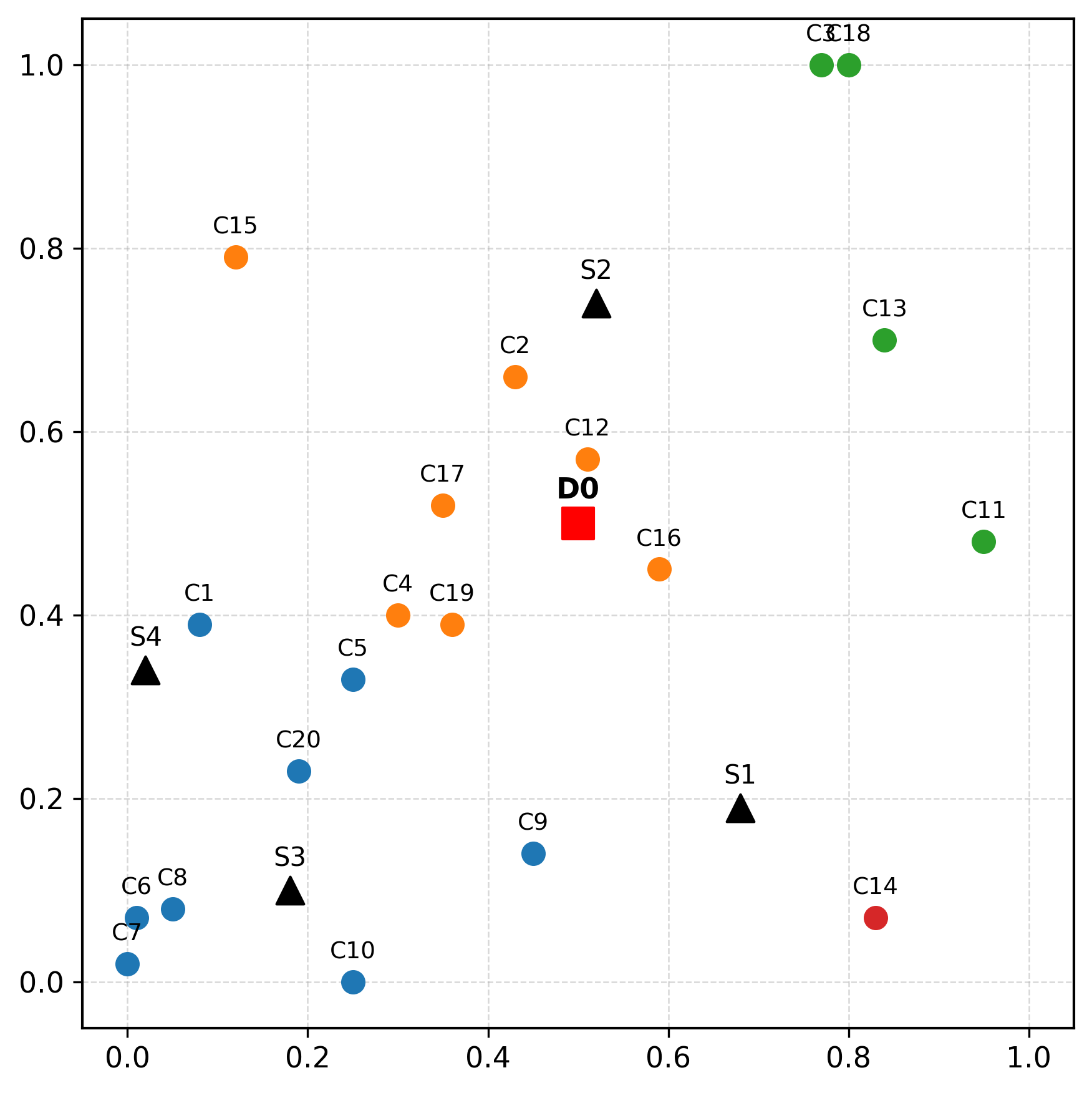}
        \put(-7,92){\textbf{(b)}}
    \end{overpic}
    \hfill
    \begin{overpic}[width=0.32\linewidth]{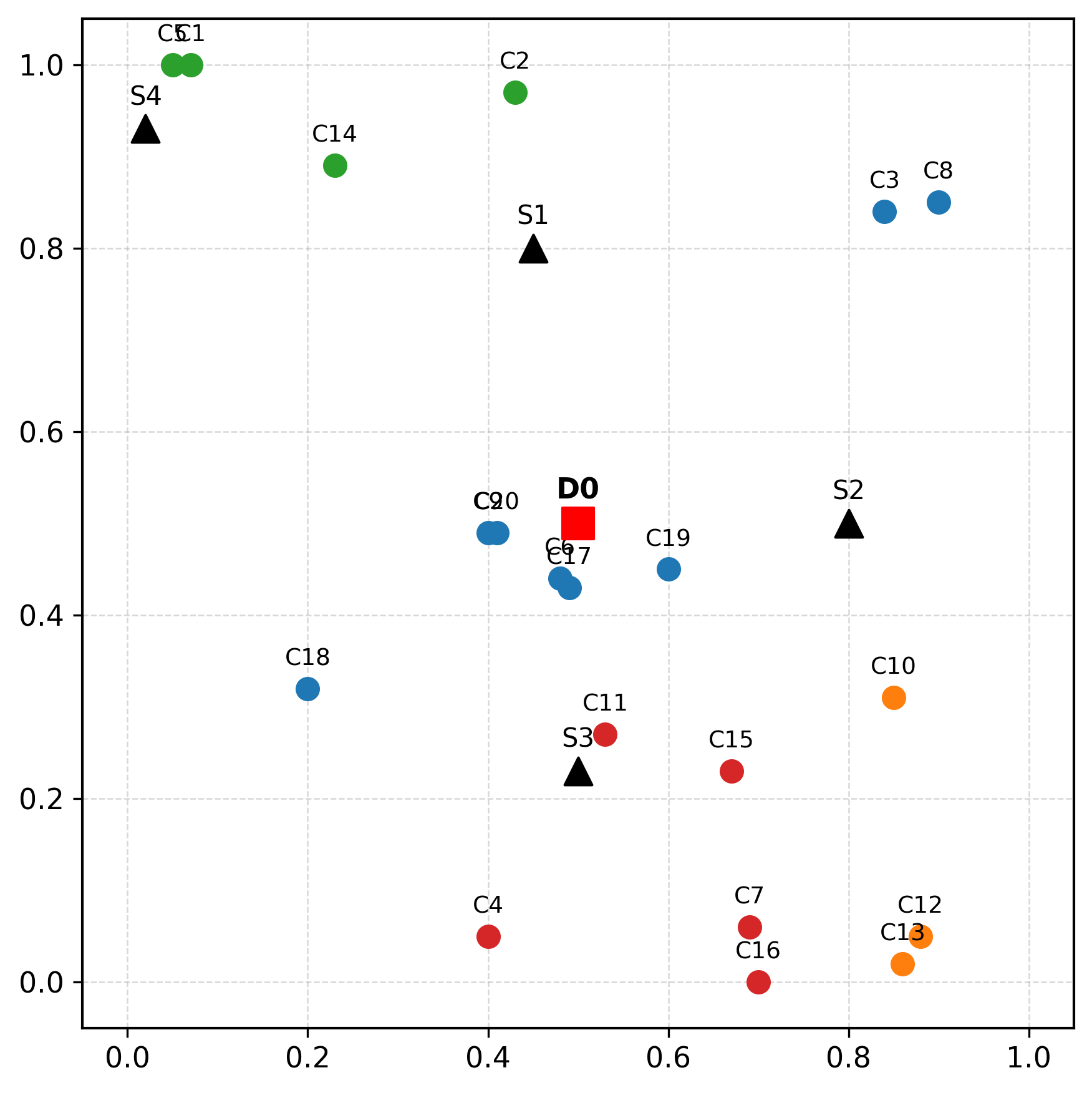}
        \put(-7,92){\textbf{(c)}}
    \end{overpic}
    
    \caption{Representative SynthCharge spatial topologies under three spatial regimes: (a) random (R), (b) clustered (C), and (c) mixed (RC). The depot is shown as a square, customers as circles, and charging stations as triangles.}
    \label{fig:topologies}
    \vspace{-10pt}
\end{figure*}

All node coordinates are generated within the unit square $[0,1]^2$. The depot location is selected according to a depot mode: \emph{center} (located at $(0.5,0.5)$), \emph{random} (sampled uniformly from $[0,1]^2$), or \emph{user-specified}. Customers are generated according to one of three spatial families inspired by the Solomon VRPTW structures \cite{Solomon1987}. For random (R) instances, each customer coordinate $\mathbf{x}_i$ is sampled independently from the uniform distribution over the unit square, i.e., $\mathbf{x}_i \sim \mathcal{U}\left([0,1]^2\right)$. A minimum pairwise separation $d_{\min}=0.04$ is enforced via rejection sampling: if a sampled point is within distance $d_{\min}$ to any previously placed customer, it is resampled. After 10 unsuccessful attempts, the point is accepted regardless of separation to guarantee termination of the procedure. For clustered (C) instances, $k$ cluster centers $\boldsymbol{\mu}_c \in [0,1]^2$ are sampled uniformly. Customer coordinates are then drawn from isotropic Gaussian distributions $\mathcal{N}(\boldsymbol{\mu}_c, \sigma^2 I)$ with any out-of-bounds coordinates clipped to the borders of the unit square $[0,1]^2$ (e.g., a generated coordinate of $(1.05, 0.4)$ is clipped to $(1.0, 0.4)$). For mixed (RC) instances, a mixing parameter $\rho \in (0,1)$ specifies the proportion of customers generated using the clustered procedure, with the remaining $1-\rho$ fraction generated using the random procedure. These spatial families, as illustrated in Fig. \ref{fig:topologies}, enable controlled variation between homogeneous and heterogeneous spatial dispersion patterns.

\subsection{Charging Infrastructure Generation}
\label{sec:stations}

Charging stations are placed using range-aware geometric rules derived from the travel range $R$. The depot always acts as a charging location. Candidate stations are generated according to two rules:

\begin{enumerate}
    \item \textbf{Customer-customer reachability.} For each customer pair $(i,j)$ with $d_{ij} > 0.8R$, a candidate station is inserted near the midpoint $(\mathbf{x}_i+\mathbf{x}_j)/2$. Specifically, we add a small bounded \emph{2D} perturbation $\boldsymbol{\epsilon}\in[-\delta,\delta]^2$ and clip the resulting coordinates to ensure they remain within the $[0,1]^2$ domain (e.g., a perturbed coordinate of $(0.5, -0.03)$ is clipped to $(0.5, 0.0)$).
    \item \textbf{Customer-depot reachability.} For any customer $i$ satisfying $d_{0i} > 0.7R$, let $s$ denote the nearest existing station. If $d_{is} > 0.5R$, an additional station is inserted at distance $0.6R$ along the ray extending from the depot to $i$. We chose $0.6R$ as a conservative distance that places the station within range of both the depot and isolated customer while providing slack relative to the travel limit $R$.
\end{enumerate}

These thresholds partition the unit square's maximum diagonal distance ($\sqrt{2} \approx 1.41$) into fractions of $R$. This ensures fundamental spatial reachability, whereas strict routing feasibility is deferred to the subsequent screening procedure.

Candidate stations are filtered to remove redundancy and overlap: stations must be separated by at least $0.3R$ and be at least $0.04$ away from any customer node. If fewer than the desired number of stations are generated, additional stations are sampled uniformly with rejection until the target count is reached. All random decisions are governed by the instance-level seed, ensuring reproducibility.

\subsection{Vehicle and Service Parameters}
\label{sec:vehparams}

Customer demands $q_i$ are sampled as fractions of vehicle load capacity $Q$ from a uniform distribution $\mathcal{U}(0.02Q, 0.30Q)$. If the aggregate demand exceeds $3Q$, the demands are proportionally rescaled so that the total demand equals $3Q$. Service durations $s_i$ are drawn independently and uniformly from the interval $[s_{\min}, s_{\max}]$. These bounds are kept fractionally small relative to average travel times so service operations do not dominate the routing schedule.

Battery capacity $B$ may be specified directly or determined adaptively. Under adaptive scaling, the maximum pairwise customer distance $d_{\max}$ is computed, and a target travel range proportional to this maximum spatial spread is defined as $R^{\star} = \kappa d_{\max}$. The effective range is then constrained to lie within a fixed normalized band $[R_{\min}, R_{\max}]$ chosen relative to the unit-square geometry. Battery capacity is finally defined as
\begin{equation*}
B = r \cdot \max\!\Bigl(R_{\min},\, \min\bigl(R_{\max},\, R^{\star}\bigr)\Bigr),
\end{equation*}
with $R_{\min}=0.15$, $R_{\max}=0.40$, and $\kappa=0.8$. This construction keeps the maximum travel range $R=B/r$ within $[0.15, 0.40]$ while adapting energy availability to realized spatial dispersion, ensuring that instances remain energy-constrained but generally reachable.

\subsection{Time Window Assignment}
\label{sec:timewindows}

Time windows are defined over a planning horizon $[0,H]$, where $H$ denotes the maximum allowable service time. A width fraction $\phi \in (0,1]$ determines a common window width $W = \phi H$ applied to all customers. Customers are indexed from 1 to $N$. Earliest service times are assigned as $e_i = (i/N)\times 0.3H$, creating a staggered temporal ordering across customers. Latest service times are set to $l_i = \min(H, e_i + W)$. If $l_i = H$, the start time $e_i$ is shifted backward so that the window width $l_i - e_i = W$ is preserved. SynthCharge supports wide ($\phi=0.8$), medium ($\phi=0.4$), and tight ($\phi=0.2$) regimes.

\subsection{Feasibility Screening Procedure}
\label{sec:certification}

SynthCharge employs a two-stage feasibility screening procedure to assign feasibility labels. The first stage removes structurally invalid configurations using deterministic checks, while the second stage optionally verifies small instances using an exact mixed integer linear program. 

\subsubsection{Stage 1: Structural Feasibility Screening}
\label{sec:screening}

The first stage applies necessary-condition checks in linear time. 
Distances are assumed to be Euclidean and symmetric, i.e., $d_{ij} = d_{ji}$. 
Three structural conditions are verified:

\begin{enumerate}
    \item \textbf{Energy reachability:} 
    Every customer $i$ must be within the maximum travel range $R$ of at least one energy-replenishing node (the depot or a charging station $s \in S$), i.e., $\min_{p \in \{0\} \cup S} d_{ip} \le R$.

    \item \textbf{Depot return consistency:}  
    Service at customer $i$ must allow return to the depot within the planning horizon, i.e., $e_i + s_i + t_{i0} \le H$.

    \item \textbf{Station accessibility:}  
    For every customer $i \in C$, there must exist at least one charging station $s \in S$
    such that $d_{is} \le R$.
\end{enumerate}

Instances failing any condition are rejected and regenerated. 
For instances with $N > 10$, only Stage~1 is applied. Note that satisfying these conditions does not ensure that a feasible route exists; it only filters out instances with trivial structural infeasibilities at negligible computational cost.

\subsubsection{Stage 2: Exact MILP Verification (Small Instances)}
\label{sec:milp}

For small instances ($N \le 10$), SynthCharge optionally invokes an exact MILP solver 
adapted from the three-index formulation of \cite{Schneider2014}. 
The formulation enforces flow conservation, time window satisfaction, and energy dynamics.

Let $x_{ij}$ be a binary variable equal to one if the vehicle travels from node $i$ to node $j$, and zero otherwise. The MILP includes the following three constraints. \textit{(1) Flow conservation (exactly-once visitation):} $\sum_{j\in V} x_{ij} = \sum_{j\in V} x_{ji} = 1$, $\forall i\in C$. These equalities ensure that each customer is entered and exited exactly once. \textit{(2) Time window satisfaction:} $b_i + s_i + t_{ij} \le b_j + M (1 - x_{ij})$, $\forall i,j\in V$, where $b_i$ denotes the service start time at node $i$ and $M$ is a sufficiently large constant. \textit{(3) Energy propagation:} $y_j \le y_i - r d_{ij} + M (1 - x_{ij})$, $\forall i,j\in V$, where $y_i$ denotes the battery level upon arrival at node $i$. The MILP verifies the existence of a route satisfying all constraints. Because exact solving is computationally expensive, this verification step is restricted to $N \le 10$.

\subsection{Generation Algorithm}
\label{sec:generation}

Algorithm~\ref{alg:generation} summarizes the procedure for generating synthetic instances. After drawing spatial coordinates and building the charging infrastructure, SynthCharge assigns demands, service times, and time windows, and then applies the two-stage feasibility screening process. 

\begin{algorithm}[t]
\caption{SynthCharge instance generation}
\label{alg:generation}
\begin{algorithmic}[1]
\Require Number of customers $N$, number of stations $S$, spatial family $f\in\{\text{R},\text{C},\text{RC}\}$, temporal regime $w\in\{\text{wide},\text{medium},\text{tight}\}$, seed, and parameter set $\Theta$ (including $k$, $\sigma$, $\rho$, $Q$, $r$, $s_{\min}$, $s_{\max}$, $H$, $\phi$)
\Ensure Feasibility\textendash filtered EVRPTW instance
\State Initialize random generator with seed
\State Generate depot location
\State Generate customer coordinates according to family $f$
\State Determine battery capacity $B$ (fixed or adaptive)
\State Construct charging infrastructure using range\textendash aware rules
\State Sample demands $q_i$ and service times $s_i$
\State Assign time windows $[e_i,l_i]$ according to regime $w$
\State Apply structural feasibility screening (Sec.~\ref{sec:screening})
\If{$N \le 10$}
    \State Perform exact MILP feasibility verification (Sec.~\ref{sec:milp})
\EndIf
\If{validation passes}
    \State Export instance and metadata
\Else
    \State Regenerate instance
\EndIf
\end{algorithmic}
\end{algorithm}

\subsection{GUI and Outputs}
SynthCharge includes an interactive GUI for configuring spatial families, temporal regimes, and energy parameters with real-time visualization. For batch generation, the generator exports instances as standardized .txt files into \texttt{feasible/} and \texttt{infeasible/} directories based on the validation outcome. Each instance is accompanied by a JSON metadata log recording the full parameter configuration and feasibility status, ensuring exact reproducibility for distribution-aware performance assessment. The complete codebase and GUI will be released as an open-source package.

\subsection{Data Reliability Metric}
The acceptance rate $\gamma$ is defined as the proportion of generated instances that pass feasibility screening; that is, $\gamma$ is the ratio of the number of validated instances to the number of generated instances. $\gamma$ provides a data reliability indicator and quantifies the interaction between spatial structure, temporal tightness, and energy scaling. For instances with $N \le 10$, $\gamma$ reflects both structural screening and exact MILP verification. For larger instances ($N > 10$), $\gamma$ is calculated using only the structural feasibility screening stage.

\section{Experimental Evaluation}
\label{sec:experiments}

This section evaluates SynthCharge as a benchmarking infrastructure.  
We examine acceptance rates across regimes and generation efficiency.  
Experiments were conducted on a workstation equipped with a 13th Gen Intel Core i7-13700K processor (16 physical cores, 24 logical processors, base frequency 3.40\,GHz) and 64\,GB of RAM.

Instance sizes included the following: 
$5\text{C}2\text{S}$, $10\text{C}3\text{S}$, $20\text{C}4\text{S}$,
$30\text{C}4\text{S}$, $40\text{C}5\text{S}$,
$50\text{C}6\text{S}$, $60\text{C}7\text{S}$,
$70\text{C}8\text{S}$, $80\text{C}9\text{S}$,
$90\text{C}10\text{S}$, and $100\text{C}12\text{S}$,
where the integer preceding $\text{C}$ denotes the number of customers and the integer preceding $\text{S}$ denotes the number of external charging stations.  
The depot always functions as an additional charging location. For each spatiotemporal regime combination (R/C/RC $\times$ wide/medium/tight), instances were generated using stochastic seeding (random seed mode). Generator configuration parameters included cluster center sampling mode set as \textit{random}, cluster spread $\sigma = 0.05$, mix ratio $\rho = 0.5$, vehicle capacity $Q = 1.5$, energy consumption rate $r = 0.25$, charging rate $g = 1.0$ (representing the amount of energy replenished per unit of time), and service times $s_i \sim \mathcal{U}(0.01, 0.03)$. Acceptance rates were computed using Stage~1 structural feasibility screening for all instances and, additionally, Stage~2 exact MILP verification for $N \le 10$.

\subsection{Acceptance Rate across Regimes}
\begin{figure}[t]
    \centering
    \includegraphics[width=\linewidth]{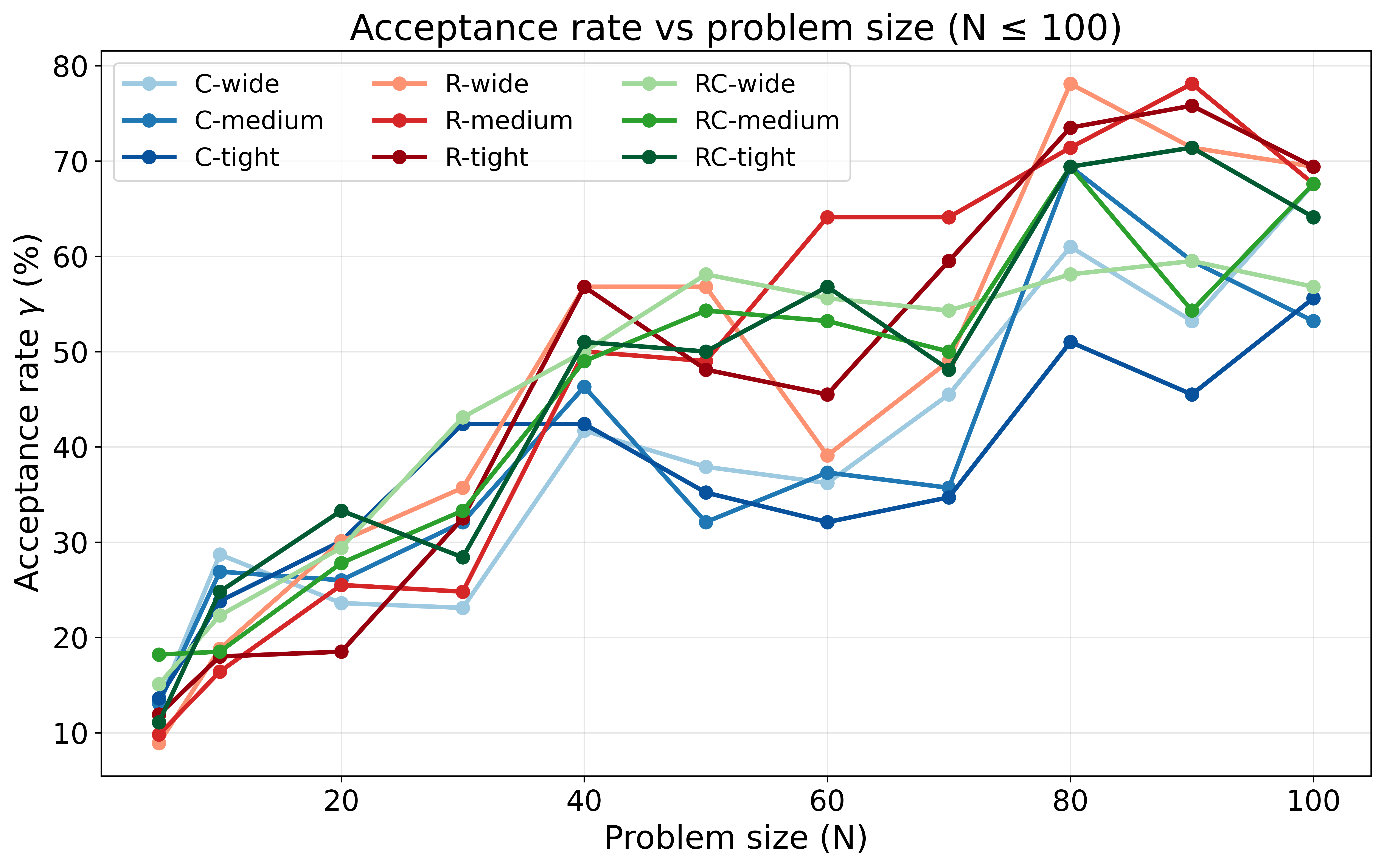}
    \caption{Acceptance rate $\gamma$ (\%) versus problem size $N$ across spatiotemporal regimes.}
    \label{fig:trends}
    \vspace{-10pt}
\end{figure}

To compute reliability metrics, we generated 2,475 feasible instances (25 per problem size across the nine spatiotemporal combinations). Table~\ref{tab:acceptance} reports the mean acceptance rate $\gamma$ (with standard deviation) across spatiotemporal regimes. Across all spatial families, tighter time windows reduce acceptance in the clustered (C) regime, whereas the random (R) and mixed (RC) regimes show weaker and sometimes non-monotonic sensitivity across wide/medium/tight settings. Figure~\ref{fig:trends} shows acceptance trends as a function of problem size $N$ for each spatiotemporal regime, demonstrating that acceptance varies with both constraint regime and scaling. Overall, results indicate that feasibility filtering is nontrivial and regime-dependent. 

\begin{table}[t]
\centering
\caption{Acceptance rate $\gamma$ (mean $\pm$ std.) across spatiotemporal regimes ($N=5$,...,$100$)}
\label{tab:acceptance}
\footnotesize
\begin{tabular}{lccc}
\toprule
\textbf{Spatial family} & \textbf{Wide} & \textbf{Medium} & \textbf{Tight} \\
\midrule
Random (R)    & 46.7 $\pm$ 22.3 & 47.4 $\pm$ 24.2 & 46.3 $\pm$ 23.2 \\
Clustered (C) & 39.3 $\pm$ 16.8 & 39.2 $\pm$ 16.4 & 37.0 $\pm$ 12.1 \\
Mixed (RC)    & 45.7 $\pm$ 16.0 & 45.1 $\pm$ 18.0 & 46.2 $\pm$ 19.6 \\
\bottomrule
\end{tabular}
\end{table}

\subsection{Generation Efficiency}
Efficient data generation is critical for training learning-based solvers. Because runtime behavior remains largely uniform across all spatiotemporal regimes, we summarize the generation times directly. The mean generation time per accepted instance starts at approximately $0.06$ seconds for $N=5$, rises to $0.10$ seconds for $N=10$, and then abruptly drops to $<0.01$ seconds for $N>10$. This sharp decrease occurs because the computationally\textendash expensive Stage~2 exact MILP verification is applied exclusively to instances with $N \le 10$. For larger instances relying solely on Stage~1 linear screening, generation time scales gradually, reaching only $0.04$ seconds for $N=100$. These results indicate that SynthCharge can serve as a high-volume instance generator while maintaining some validation transparency.

\begin{figure}[t]
    \centering
    \includegraphics[width=0.98\linewidth]{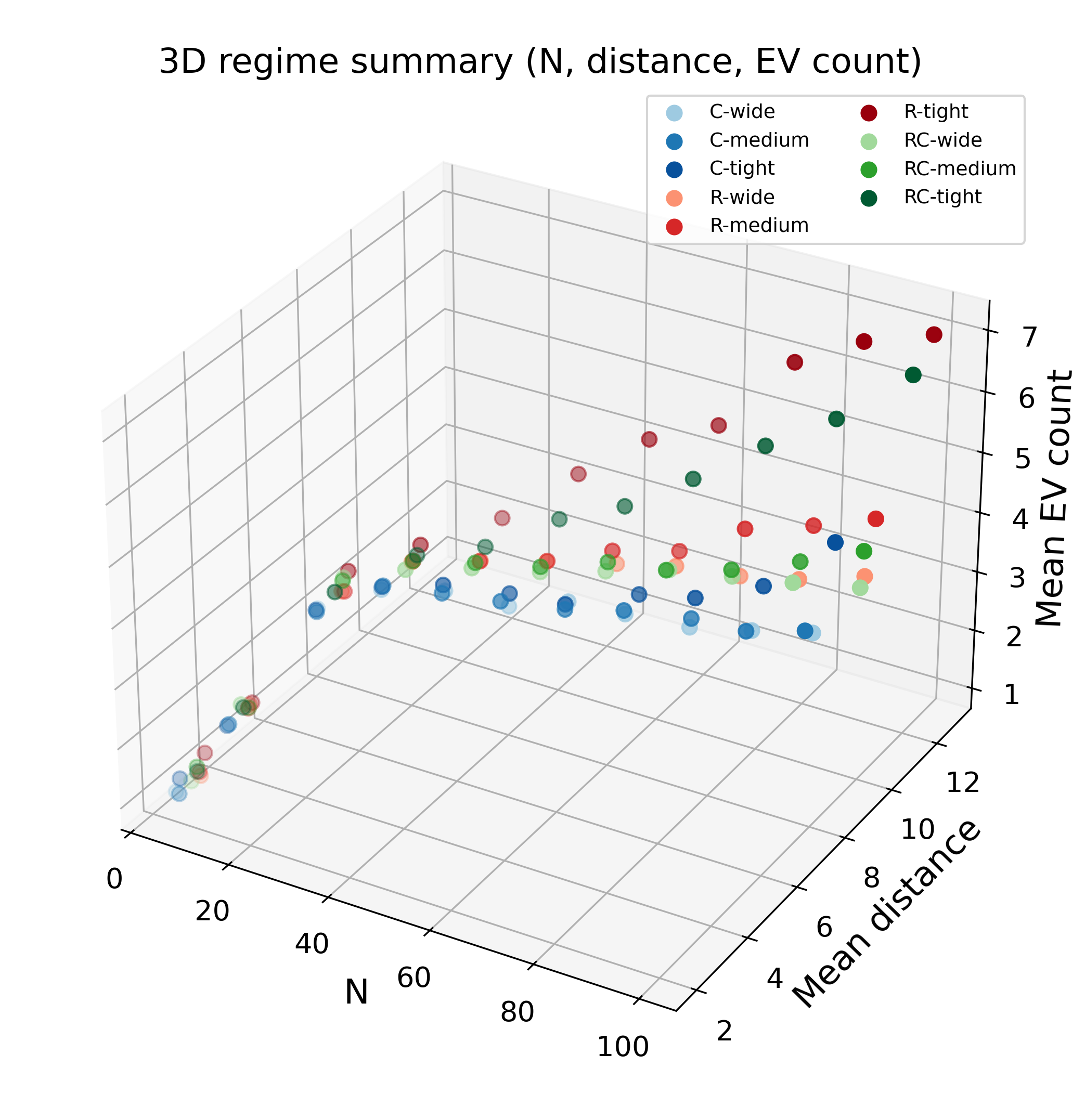}
    \caption{Problem size versus mean traveled distance and mean EV count for random (R), clustered (C), and mixed (RC) regimes under a baseline metaheuristic.  The trends illustrate how temporal tightening increases fleet requirements.}
    \vspace{-12pt}
    \label{fig:evcount}
    
\end{figure}

\subsection{Empirical Feasibility Evaluation and Resource Utilization}
To verify practical solvability, we evaluated  all generated instances across all spatiotemporal regimes ($N=5, \dots, 100$) using a baseline variable neighborhood search/tabu search metaheuristic \cite{Schneider2014}. The algorithm successfully produced feasible routes for $100\%$ of these instances, confirming that Stage~\ref{sec:screening} structural screening strongly correlates with actual routing feasibility. This evaluation also reinforces the operational impact of tightening constraints. As illustrated in Figure~\ref{fig:evcount}, mean travel distance and fleet size increase monotonically as time windows tighten and spatial topologies become more dispersed, with the most pronounced growth occurring in random and mixed configurations.

\section{Conclusions}
\label{sec:conclusion}

This paper presents SynthCharge, a configurable generator for creating benchmark instances of EVRPTW. SynthCharge produces synthetic instances with controlled spatial layouts, customer scales, time-window settings, and energy constraints, while filtering out structurally infeasible cases through an explicit screening procedure. By allowing systematic variation of these problem characteristics, SynthCharge enables rigorous and reproducible evaluation of both optimization- and learning-based routing methods. This is particularly relevant for learning-based EVRPTW methods, where performance can degrade under distribution shift and reproducible comparison increasingly depends on explicit generation assumptions and feasibility checks. Several limitations define the current scope. Exact MILP verification is restricted to small instances ($N \le 10$), and structural screening remains necessary but not sufficient for global feasibility. Instances are generated in a Euclidean unit-square domain with deterministic travel times and a single-depot setting; non-Euclidean road geometries, time-dependent travel, stochastic dynamics, and multi-depot configurations are not modeled. Charging behavior assumes linear energy consumption and full recharge decisions, excluding nonlinear charging curves and partial recharging policies. These design choices prioritize interpretability, controlled distributional variation, and computational efficiency within a benchmarking-oriented infrastructure. Future work includes extending the generator to multi-depot settings, nonlinear and partial charging models, time-dependent travel, and integration with real road network geometries. We also plan to release standardized datasets and benchmarking protocols to support reproducible EVRP evaluation.  


\bibliographystyle{IEEEtran}

\small 
\begin{thebibliography}{21}
\setlength{\itemsep}{0pt} 
\setlength{\parskip}{0pt}


\bibitem{Schneider2014}
M.~Schneider, A.~Stenger, and D.~Goeke, ``The electric vehicle-routing problem with time windows and recharging stations,'' \emph{Transportation Science}, vol.~48, no.~4, pp.~500--520, 2014.

\bibitem{Erdogan2012}
S.~Erdogan and E.~Miller-Hooks, ``A green vehicle routing problem,'' \emph{Transportation Research Part~E: Logistics and Transportation Review}, vol.~48, no.~1, pp.~100--114, 2012.

\bibitem{Desaulniers2016}
G.~Desaulniers, F.~Errico, S.~Irnich, and M.~Schneider, ``Exact algorithms for electric vehicle-routing problems with time windows,'' \emph{Operations Research}, vol.~64, no.~6, pp.~1388--1405, 2016.

\bibitem{Hiermann2016}
G.~Hiermann, J.~Puchinger, S.~Ropke, and R.~F.~Hartl, ``The electric fleet size and mix vehicle routing problem with time windows and recharging stations,'' \emph{European Journal of Operational Research}, vol.~252, no.~3, pp.~995--1018, 2016.

\bibitem{Keskin2016}
M.~Keskin and B.~Çatay, ``Partial recharge strategies for the electric vehicle routing problem with time windows,'' \emph{Transportation Research Part~C: Emerging Technologies}, vol.~65, pp.~111--127, 2016.

\bibitem{goeke2019}
D.~Goeke and M.~Schneider, ``Routing a mixed fleet of electric and conventional vehicles,'' \emph{European Journal of Operational Research}, vol.~245, no.~1, pp.~81--99, 2015.

\bibitem{nazari2018reinforcement}
M.~Nazari, A.~Oroojlooy, L.~V.~Snyder, and M.~Takác, ``Reinforcement learning for solving the vehicle routing problem,'' in \emph{Advances in Neural Information Processing Systems}, vol.~31, 2018.

\bibitem{kool2019attention}
W.~Kool, H.~van Hoof, and M.~Welling, ``Attention, learn to solve routing problems!'' in \emph{Proc. International Conference on Learning Representations}, 2019.


\bibitem{DeAndoin2023EVCRP}
M.~G.~De Andoin, A.~Bottarelli, S.~Schmitt, I.~Oregi, P.~Hauke, and M.~Sanz,
``Formulation of the Electric Vehicle Charging and Routing Problem for a Hybrid Quantum-Classical Search Space Reduction Heuristic,''
in \emph{2023 IEEE 26th International Conference on Intelligent Transportation Systems (ITSC)},
pp.~5318--5323, 2023, doi:10.1109/ITSC57777.2023.10421787.

\bibitem{Peng2024RobustBET}
D.~Peng, M.~J.~Barth, and K.~Boriboonsomsin,
``Addressing the Robust Battery Electric Truck Dispatching Problem with Backhauls and Time Windows Under Travel Time Uncertainty,''
in \emph{2024 IEEE 27th International Conference on Intelligent Transportation Systems (ITSC)},
pp.~84--89, 2024, doi:10.1109/ITSC58415.2024.10919697.

\bibitem{DeNunzio2020ITSC}
G.~De~Nunzio, I.~Ben~Gharbia, and A.~Sciarretta,
``A Time- and Energy-Optimal Routing Strategy for Electric Vehicles with Charging Constraints,''
in \emph{2020 IEEE 23rd International Conference on Intelligent Transportation Systems (ITSC)},
pp.~1--8, 2020, doi:10.1109/ITSC45102.2020.9294622.

\bibitem{Jiang2023Shift}
Y.~Jiang, Y.~Wu, Z.~Zheng, and Z.~Wang, ``Ensemble-based deep reinforcement learning for vehicle routing problems under distribution shift,'' in \emph{Advances in Neural Inf. Processing Systems}, vol.~36, 2023.

\bibitem{Zhou2023Omni}
J.~Zhou \emph{et~al.}, ``Towards omni-generalizable neural methods for vehicle routing problems,'' in \emph{Proceedings of the International Conference on Machine Learning}, 2023.

\bibitem{Luo2025Tightness}
F.~Luo, Y.~Wu, Z.~Zheng, and Z.~Wang, ``Rethinking neural combinatorial optimization for VRPs with different constraint tightness degrees,'' \emph{arXiv preprint arXiv:2505.24627}, 2025.


\bibitem{Daysalilar2026CB}
M.~Daysalilar, F.~Uyguroglu, G.~Nicolosi, and A.~Meyers, ``A curriculum-based deep reinforcement learning framework for the electric vehicle routing problem,'' \emph{arXiv preprint arXiv:2601.15038}, 2026.

\bibitem{Solomon1987}
M.~M.~Solomon, ``Algorithms for the vehicle routing and scheduling problems with time window constraints,'' \emph{Operations Research}, vol.~35, no.~2, pp.~254--265, 1987.

\bibitem{uchoa2017}
E.~Uchoa, D.~Pecin, A.~Pessoa, M.~Poggi, T.~Vidal, and A.~Subramanian, ``New benchmark instances for the capacitated vehicle routing problem,'' \emph{European Journal of Opr. Research}, vol.~257, no.~3, pp.~845--858, 2017.

\bibitem{vrprep}
J.~E.~Mendoza, C.~Arnold, D.~Sörensen, and K.~Tierney, ``VRP-REP: The vehicle routing community repository,'' 2014. 


\bibitem{Nigam2023ITSC}
A.~Nigam and S.~Srivastava, ``Generating realistic synthetic traffic data using conditional tabular GANs for ITS,'' in \emph{Proc. IEEE International Conference on Intelligent Transportation Systems}, 2023.

\bibitem{Gebru2021Datasheets}
T.~Gebru \emph{et~al.}, ``Datasheets for datasets,'' \emph{Communications of the ACM}, vol.~64, no.~12, pp.~86--92, 2021.


\bibitem{Daysalilar2023Heterogeneous}
M.~Daysalilar, C.~B.~Chen, and M.~Erkoc, ``Electric vehicle routing problems for heterogeneous fleets with partial recharging, diverse charger types, and soft time windows,'' in \emph{Proc. IISE Annual Conference}, 2023, pp.~1--6.  doi:10.21872/2023IISE\_2943.


\end{thebibliography}

\end{document}